\pdfoutput=1
\documentclass[runningheads]{llncs}
\usepackage{multirow}
\usepackage{graphicx}
\usepackage{tabularx}
\usepackage{hyperref}
\usepackage{xurl}
\title{K-12BERT: BERT for K-12 education}
\author{Vasu Goel\inst{1} \and Dhruv Sahnan\inst{1} \and Venktesh V\inst{1} \and Gaurav Sharma\inst{2} \and Deep Dwivedi\inst{1} \and
Mukesh Mohania \inst{1}}
\tocauthor{Vasu ~ Goel,  Dhruv ~ Sahnan, venktesh ~ V, Gaurav ~ Sharma,  Deep ~ Dwivedi , Mukesh  ~ Mohania}
\date{}
\toctitle{K-12BERT: BERT for K-12 education}
\authorrunning{Vasu Goel et al.}
\def\MODELNAME{\texttt{K-12Bert}}
\def\MODELSENTNAME{\texttt{K-12Sent\_BERT}}
\newcommand{\modelName}{\MODELNAME}
\newcommand{\modelSentName}{\MODELSENTNAME}
\institute{
    Indraprastha Institute of Information Technology, Delhi \\ \email{\{vasu18322,dhruv18230,venkteshv,mukesh, deepd\}@iiitd.ac.in}
    \and 
    Extramarks Education pvt. ltd.\\ 
    \email{gaurav.sharma@extramarks.com}
}
\begin{document}

\maketitle

\begin{abstract}
Online education platforms are powered by various NLP pipelines, which utilize models like BERT to aid in content curation. Since the inception of the pre-trained language models like BERT, there have also been many efforts toward adapting these pre-trained models to specific domains. However, there has not been a model specifically adapted for the  education domain (particularly K-12) across subjects to the best of our knowledge. In this work, we propose to train a language model on a corpus of data curated by us across multiple subjects from various sources for K-12 education. We also evaluate our model, \modelName{}, on downstream tasks like hierarchical taxonomy tagging.

\end{abstract}
\section{Introduction}
The pre-trained language models like BERT\cite{bert} have made considerable advancements in many NLP tasks. However, these models are trained on general domain text like Wikipedia and Book Corpus and are not adapted to the vocabulary of the target domain. Several works have addressed this by training domain-specific language models like PubMedBERT\cite{Gu_2022}, SciBERT\cite{SciBERT}, BioBERT\cite{BioBERT} for biomedical NLP. Following their success over vanilla pre-trained models, several other works like TravelBERT\cite{TravelBERT}, PatentBERT\cite{PatentBERT}, FinBERT\cite{FinBERT} have adapted domain-specific pre-training for the respective domains. The domain-specific pre-training can be performed in two ways: the continued pre-training approach
or pre-training from scratch. For instance, BioBERT and SciBERT demonstrate that when the corpus is small leveraging pre-trained models to continue training on domain-specific corpus leads to an increase in performance on downstream in-domain tasks. Additionally, works like PubMedBERT demonstrate that pre-training from scratch leads to gains on downstream tasks when in domain corpus is abundant.

\begin{table}[hbt!]
\small
\caption{Composition of our corpus.}
\label{tab:dataset}
\begin{tabular}{|p{0.44\textwidth}|p{0.24\textwidth}|p{0.25\textwidth}|}

 \hline
 \textbf{Source} & \textbf{Content} & \textbf{\# sentences} \\ 
 \hline
 NCERT(India) & P, C, B, SS  & 15K \\ 
 \hline
 Siyavulla.com (International) & H, L   & 2K \\ 
 \hline
 OpenStax.org (USA) & P, C, B  & 4K \\ 
 \hline
 Learncbse.in (India)    & P, C, B, SS  & 19K \\ 
 \hline
 CK-12.org (USA) &  P, C, B, L, H, E   & 14K \\ 
 \hline
 KhanAcademy.org (USA) & P, C, B, SS  & 282K \\ 
 \hline
 Extramarks.com (India) & P, C, B, H, SS  & 120K \\ 
 \hline
\end{tabular}
Physics (P), Chemistry (C), Biology (B), Social studies (SS), Physical science (H), Life science (L), Earth science (E)
\end{table}

In several instances, one might posit that the general domain text may overlap with the education vocabulary. However, the general domain text may contain advanced terms while lacking academic concepts, which are crucial for students to understand and achieve the learning objectives. This paper proposes to train the BERT model on a corpus of text collected from various sources for the K-12 education system across different geographic regions. We perform continued pre-training as the corpus is not abundant compared to other domains. In summary, the following are the core contributions of our work:
\begin{itemize}
    \item We release a corpus for the K-12 education system as shown in Table \ref{tab:dataset}.
    \item We perform continued pre-training of BERT on the K-12 corpus and evaluate on downstream tasks.
    \item Code and data are at \url{https://github.com/ADS-AI/K12-Bert-AIED-2022}
\end{itemize}
\section{Methodology}

\subsection{Dataset}
We curate our dataset from multiple online learning platforms that provide open access for research purposes. To the best of our knowledge, the dataset curated is the first of its kind due to the lack of a corpus of K-12 learning content suitable for language model training. Table \ref{tab:dataset} describes the details of the datasets collected. The data collected ranges across different regions like the USA, India, and South Africa to avoid regional bias in the dataset.
\begin{table}[hbt!]
\small
\caption{\label{tab:results}Performance comparison (Recall@K) of \modelName with other baselines.}
\begin{tabular}{|p{0.1\textwidth}|p{0.45\textwidth}|p{0.09\textwidth}|p{0.09\textwidth}|p{0.09\textwidth}|p{0.09\textwidth}|}
\hline
\textbf{Dataset}                           & \textbf{Model}                  & \textbf{R@5} & \textbf{R@10} & \textbf{R@15} & \textbf{R@20} \\ \hline
\multicolumn{1}{|l|}{\multirow{6}{*}{ARC}} & BERT+USE                & 0.67         & 0.81          & 0.86          & 0.89          \\ \cline{2-6} 
\multicolumn{1}{|l|}{}                     & BERT+Sent\_BERT         & 0.65         & 0.77          & 0.84          & 0.88          \\ \cline{2-6} 
\multicolumn{1}{|l|}{}                     & BERT+\modelSentName{} & 0.68        & 0.81          & 0.87          & 0.90         \\ \cline{2-6} 
\multicolumn{1}{|l|}{}                     & \modelName+USE            & 0.65         & 0.78          & 0.85          & 0.88          \\ \cline{2-6} 
\multicolumn{1}{|l|}{}                     & \modelName+Sent\_BERT & 0.68         & \textbf{0.82}          & \textbf{0.87}          & 0.90         \\ \cline{2-6} 
\multicolumn{1}{|l|}{}                     & \modelName+\modelSentName & \textbf{0.68}        & 0.81   & 0.86   & \textbf{0.90}      \\ \hline
\multicolumn{1}{|l|}{\multirow{6}{*}{QC-Science}}  & BERT+USE                & 0.86         & 0.92          & 0.95          & 0.96          \\ \cline{2-6} 
\multicolumn{1}{|l|}{}                     & BERT+Sent\_BERT         & 0.85         & 0.93          & 0.95          & 0.97          \\ \cline{2-6}
\multicolumn{1}{|l|}{}                     & BERT+\modelSentName             & 0.88         & \textbf{0.94}         & 0.96          & 0.97         \\ \cline{2-6} 
\multicolumn{1}{|l|}{}                     & \modelName+USE             & 0.84         & 0.91          & 0.94          & 0.96          \\ \cline{2-6} 
\multicolumn{1}{|l|}{}                     & \modelName+Sent\_BERT            & 0.88         & 0.93          & 0.95          & 0.97          \\ \cline{2-6} 
\multicolumn{1}{|l|}{}                     & \modelName+\modelSentName & \textbf{0.88}         & 0.94          & \textbf{0.96}          & \textbf{0.97}          \\ \hline
\end{tabular}

\end{table}
The data collected is as follows:
\begin{itemize}
    \item For \textit{NCERT (K-12) (India)} we used pdfminer\footnote{https://pypi.org/project/pdfminer2/}, a python library for extracting information from NCERT PDFs\footnote{https://ncert.nic.in/textbook.php}.
    \item For \textit{Siyavulla, OpenStax, LearnCBSE, CK-12} we systematically scraped the webpages which contained information and picked out the chunks which contained meaningful information. 
    Then we used the HTML parser present with BeautifulSoup4\footnote{\url{https://www.crummy.com/software/BeautifulSoup/bs4}} to break down the document into retrievable components and extract information from the paragraph tags.
    \item We accessed \textit{Khan Academy transcripts} using the official APIs\footnote{\url{https://github.com/Khan/khan-api}}. Khan academy transcripts had informal language, which added to the noise since they are made for an online video educational setup which was filtered out.
    \item The \textit{Extramarks (EM)} transcripts were made available by Extramarks in .docx format and had content in Hindi and English. We filtered out the text containing Roman Hindi characters by comparing their unicode values. Using pyenchant\footnote{https://pypi.org/project/pyenchant/} library, we run spellcheck on the words and maintain a count of approved words by the spell checker. Finally, we extract sentences that have more approved words than rejected words.
\end{itemize} 

\subsection{Continued pre-training} 
Due to the constraint on the data available in this domain, we decided to continue pre-training. For our current experiment, we do not update the existing vocabulary. Using the existing BERT vocabulary allows the model to capture diverse information and be well suited for education-related tasks.
The data that we scraped had discontinuity in sentences due to the scraping mechanism, which is a downside for NSP (Next Sentence Prediction) objective. Hence, for training \modelName\ , we use only the MLM (Masked Language Modeling) objective. To make the training resource-efficient, we utilize training techniques like Gradient Checkpointing, Gradient Accumulation, and mixed-precision training, proven to save GPU memory and speed up training. We continue the pre-training for 10 epochs over a batch size of 32 and gradient accumulation step size of 4. This setup allowed us to train our setup over 2 GPUs of 16GB memory each. We performed extensive experiments by training our model over different combinations of curated datasets. We achieved the best performance when the model was trained over Siyavulla, OpenStax, LearnCBSE, Ck-12.org, and EM transcripts.

\section{Results}
To validate the training of \modelName{} we test it on the automated question tagging task for education domain. We evaluate our model on the task presented in \cite{TagRec} of tagging learning content like questions to a hierarchical learning taxonomy of form subject - chapter - topic. We use the official code provided by the authors and replace the models and sentence encoder, keeping all other settings the same. The results of \modelName{} and previous best baselines are listed in Table \ref{tab:results}. We use state of the art models for generating contextualized sentence embeddings like Universal Sentence Encoder (USE)\cite{cer2018universal} and Sentence-BERT\footnote{https://www.sbert.net} (Sent\_BERT) to generate embeddings for the taxonomy. We noticed that \modelName+\{USE,Sent\_BERT\} wasn't performing as well as the vanilla BERT baseline. We believe the reason behind that could be since BERT, USE and SentBERT are trained on a general dataset, where as \modelName{} is exposed to more data pertaining educational domain, the embeddings generated are farther away in the vector space. So, in order to validate our hypothesis, we trained \modelSentName{}, a sentence BERT model finetuned for educational domain. We see that with using \modelName{} with \modelSentName\ we were able to outperform the previous baselines by 2\% for the QC-Science dataset and 1\% for ARC. We strongly attribute these results to the domain specific training of the models.


%
%
%
\bibliographystyle{splncs04}
\bibliography{refs}

\begin{thebibliography}{1}
\providecommand{\url}[1]{\texttt{#1}}
\providecommand{\urlprefix}{URL }
\providecommand{\doi}[1]{https://doi.org/#1}

\bibitem{FinBERT}
Araci, D.: Finbert: Financial sentiment analysis with pre-trained language
  models (2019), \url{https://arxiv.org/abs/1908.10063}

\bibitem{SciBERT}
Beltagy, I., Lo, K., Cohan, A.: Scibert: A pretrained language model for
  scientific text  (2019), \url{https://arxiv.org/abs/1903.10676}

\bibitem{cer2018universal}
Cer, D., Yang, Y., yi~Kong, S., Hua, N., Limtiaco, N., John, R.S., Constant,
  N., Guajardo-Cespedes, M., Yuan, S., Tar, C., Sung, Y.H., Strope, B.,
  Kurzweil, R.: Universal sentence encoder (2018)

\bibitem{bert}
Devlin, J., Chang, M.W., Lee, K., Toutanova, K.: Bert: Pre-training of deep
  bidirectional transformers for language understanding (2018),
  \url{https://arxiv.org/abs/1810.04805}

\bibitem{Gu_2022}
Gu, Y., Tinn, R., Cheng, H., Lucas, M., Usuyama, N., Liu, X., Naumann, T., Gao,
  J., Poon, H.: Domain-specific language model pretraining for biomedical
  natural language processing. {ACM} Transactions on Computing for Healthcare
  \textbf{3}(1),  1--23 (jan 2022), \url{https://doi.org/10.1145%2F3458754}

\bibitem{PatentBERT}
Lee, J.S., Hsiang, J.: Patentbert: Patent classification with fine-tuning a
  pre-trained bert model (2019), \url{https://arxiv.org/abs/1906.02124}

\bibitem{BioBERT}
Lee, J., Yoon, W., Kim, S., Kim, D., Kim, S., So, C.H., Kang, J.: {BioBERT}: a
  pre-trained biomedical language representation model for biomedical text
  mining. Bioinformatics  (sep 2019),
  \url{https://doi.org/10.1093%2Fbioinformatics%2Fbtz682}

\bibitem{TagRec}
V, V., Mohania, M., Goyal, V.: Tagrec: Automated tagging of questions with
  hierarchical learning taxonomy (2021), \url{https://arxiv.org/abs/2107.10649}

\bibitem{TravelBERT}
Zhu, H., Peng, H., Lyu, Z., Hou, L., Li, J., Xiao, J.: Travelbert: Pre-training
  language model incorporating domain-specific heterogeneous knowledge into a
  unified representation (2021), \url{https://arxiv.org/abs/2109.01048}

\end{thebibliography}
%





\end{document}